\pdfoutput=1

\documentclass[11pt]{article}

\usepackage{ACL}

\usepackage{times}
\usepackage{latexsym}

\usepackage[T1]{fontenc}

\usepackage[utf8]{inputenc}

\usepackage{microtype}

\usepackage{inconsolata}

\usepackage{lipsum}  
\usepackage{tabularx}
\usepackage{booktabs}

\usepackage{microtype}
\usepackage{type1cm}
\usepackage[american]{babel}
\usepackage{amssymb}
\usepackage{multirow}
\usepackage{xspace}
\DeclareMathAlphabet{\mathcalbf}{OMS}{pzc}{b}{n}
\usepackage{enumitem}
\usepackage{colortbl}
\usepackage{enumitem}

\RequirePackage{color}
\definecolor{darkgray}{gray}{0.40}
\definecolor{mediumgray}{gray}{0.60}
\definecolor{lightgray}{gray}{0.95}
\definecolor{ultralightgray}{gray}{0.98}
\definecolor{forestgreen}{rgb}{0.133, 0.545, 0.133}
\definecolor{orange}{rgb}{1, 0.86, 0.74}
\definecolor{lightergreen}{rgb}{0.95, 1, 0.88}

\usepackage{graphicx}
\DeclareGraphicsExtensions{.pdf,.ai,.jpg,.png}
\setkeys{Gin}{pagebox=artbox}
\graphicspath{{./naacl24-learner-corpus-figures/}}

\newcommand{\bsfigure}[3][]{%
    \begin{figure}[t]
        \centering
        \includegraphics[#1]{#2}
        \caption{#3}\label{#2}%
    \end{figure}
}

\newcommand{\hwfigure}[3][t!]{%
    \begin{figure*}[#1]
        \centering
        \includegraphics[scale=1.0]{#2}
        \caption{#3}\label{#2}%
    \end{figure*}
}

\RequirePackage{type1cm}
\RequirePackage{color}
\RequirePackage{soul}
\setstcolor{blue}
\definecolor{violet}{rgb}{0.5,0.0,0.5}

\newsavebox\bscombox
\newcommand{\bscom}[3][]{%
    \sbox{\bscombox}{\fontsize{8}{9}\selectfont#1#2#3}
    \noindent
    \st{#2}{\selectfont
        \color{blue}#3\ifx\\#1\\\else{\fontsize{8}{9}\selectfont\color{violet}[#1]}\fi
    }
}

\begin{document}


\title{A School Student Essay Corpus for \\ Analyzing Interactions of Argumentative Structure and Quality}

\author{Maja Stahl\textsuperscript{1}, Nadine Michel\textsuperscript{2}, Sebastian Kilsbach\textsuperscript{2}, \\ 
	{\bf Julian Schmidtke\textsuperscript{1}, Sara Rezat\textsuperscript{2}, and Henning Wachsmuth\textsuperscript{1}} \\
	\textsuperscript{1}Leibniz University Hannover, Institute of Artificial Intelligence \\
	\textsuperscript{2}Paderborn University, Institute for German Language and Comparative Literature \\
	\small{\tt \{m.stahl,h.wachsmuth\}@ai.uni-hannover.de, julian.schmidtke@stud.uni-hannover.de} \\ 
	\small{\tt \{nadine.michel, sebastian.kilsbach, sara.rezat\}@uni-paderborn.de} \\ 
}


\maketitle

\begin{abstract}

Learning argumentative writing is challenging. Besides writing fundamentals such as syntax and grammar, learners must select and arrange argument components meaningfully to create high-quality essays. To support argumentative writing computationally, one step is to mine the argumentative structure. When combined with automatic essay scoring, interactions of the argumentative structure and quality scores can be exploited for comprehensive writing support. Although studies have shown the usefulness of using information about the argumentative structure for essay scoring, no argument mining corpus with ground-truth essay quality annotations has been published yet. Moreover, none of the existing corpora contain essays written by school students specifically. To fill this research gap, we present a German corpus of 1,320 essays from school students of two age groups. Each essay has been manually annotated for argumentative structure and  quality on multiple levels of granularity. We propose baseline approaches to argument mining and essay scoring, and we analyze interactions between both tasks, thereby laying the ground for quality-oriented argumentative writing support.

\end{abstract}

\newcommand{\bspubtag}{
	\vspace*{-18.75cm}\hspace*{-0.5cm}
	{\fontsize{6}{8}\selectfont%
		\renewcommand{\arraystretch}{0.9}
		\begin{tabular}{l}
			Accepted to the North American Chapter of the Association for Computational Linguistics: NAACL 2024
		\end{tabular}
}}

\bspubtag
\vspace*{17.5cm}\hspace*{0.5cm}

\section{Introduction}
\label{sec:introduction}

Writing argumentative texts, in particular argumentative essays, constitutes an essential part of school students' writing education. However, learning to write arguments of high quality can be challenging \cite{zhu-2001-performing,ferretti-etal-2007-improving,peloghitis-2017-difficulties,alexander-etal-2023-confronting}. It requires various skills, from writing fundamentals, such as syntax and grammar, to argumentation-specific skills, such as meaningfully organizing and structuring arguments and counter-considerations \cite{rezat-2011-schriftliches}. This takes time and effort to master \cite{kakandee-kaur-2014-argumentative,dang-etal-2020-study}. Given teachers' limited time to give students feedback on their writing, automatic argumentative writing support could benefit students as it offers guidance at their own pace and convenience \cite{wambsganss-etal-2022-alen}.

\bsfigure{example-essay3}{Exemplary annotated school student essay on the use of school funding, taken from our corpus. The text is from the FD-LEX corpus \cite{becker-mrotzek-grabrowski-2018-textkorpus}, translated from German for display.}

Argumentative writing support systems employ argument mining to analyze input texts \cite{stab-2017-argumentative,wambsganss-niklaus-2022-modeling,weber-etal-2023-structured}, that is, computational methods that identify argumentative components and their relations. Common components are {\it major claim} (main standpoint of the text, also known as {\it thesis}), {\it claim} (controversial statement), and {\it premise} (reason for justifying or refuting the claim) along with their argumentative relations {\it support} and {\it attack}. This knowledge enables the systems to give feedback on the structure of a text, e.g., by highlighting unwarranted claims \cite{stab-gurevych-2017-parsing}, or by analyzing the number of argumentative components quantitatively \cite{stab-gurevych-2017-parsing,wambsganss-niklaus-2022-modeling,weber-etal-2023-structured}.

Unlike argument mining, automated essay scoring explicitly evaluates essay quality, either holistically \cite{uto-etal-2020-neural,yang-etal-2020-enhancing,wang-etal-2023-aggregating} or specific linguistic aspects, such as coherence \cite{li-etal-2018-coherence,farag-etal-2018-neural}, grammar \cite{tambe-etal-2022-automated}, and organization \cite{persing-etal-2010-organization,rahimi-etal-2015-incorporating}. 
Combining argument mining with essay scoring may enable support systems to give students comprehensive feedback on their writing. In addition, it helps identify how different argumentative structures influence the overall essay quality and which structures are common for different levels of quality \cite{wachsmuth-etal-2016-using}. However, student essay corpora for argument mining are scarce and do not include ground-truth essay quality annotations \cite{stab-gurevych-2017-parsing}. Moreover, no corpus with structure annotations for essays written by school students has been published yet.

To fill this research gap, we present a German corpus of 1,320 school student essays with manual annotations for argumentative structure and essay quality. The essays have been systematically selected from an existing corpus \cite{becker-mrotzek-grabrowski-2018-textkorpus}, equally distributed over two age groups (fifth-graders and ninth-graders) and binary genders, three per student. We present an extensive annotation scheme focused on school student essays that covers argumentative structure on four levels of granularity as well as five essay quality aspects, as shown in Figure~\ref{example-essay3}. To achieve consistent annotations, we developed annotation guidelines in close dialogue with our expert annotators from the field of language education. This led to high agreement between the annotations.

Our analyses of the corpus provide various insights into the correlation between the different levels of argumentative structure and essay quality, as well as the interaction between these two types of annotation. We experiment with fine-tuned transformers and adapters as baseline approaches to mining argumentative structure and scoring essay quality. Moreover, we demonstrate that the information on argumentative structure helps predicting the essay quality, which is in line with what previous studies showed on other corpora \cite{wachsmuth-etal-2016-using,beigman-klebanov-etal-2016-argumentation,nguyen-litman-2018-argument}. This result underlines the usefulness of our corpus annotations for quality-oriented argumentative writing support.

More explicitly, this work aims to answer (i) how the argumentative structure and essay quality of school student essays can be modeled, (ii) how different levels of argumentative structure and essay quality correlate for school student essays, and (iii) how this correlation can be exploited to automatically score the essay quality.

Altogether, this paper's main contributions are:
\begin{itemize}[itemsep=0pt]
\item A corpus for studying argumentative structure and essay quality on school student essays
\item Empirical insights into the interactions of argumentative structure and essay quality
\item Baseline approaches to argument mining and essay scoring%
\footnote{Our corpus and experiment code can be found under: \url{https://github.com/webis-de/NAACL-24}.}
\end{itemize}

\section{Related Work}
\label{sec:related-work}

Argumentative writing is a key capability that is taught in school across age groups and disciplines \cite{becker-etal-2010-argumentatives,rezat-2011-schriftliches}. A common educational form of argumentative text is the essay, where school students should introduce a thesis, to which they provide pro and con arguments, and finally conclude \cite{townsend-etal-1993-effects,schroeter-2021-linguistische}. The components of an argumentative text take on different roles (e.g., \emph{claim} or \emph{premises}), and they may operationalize different actions (e.g., \emph{conceding} or \emph{reasoning}) \cite{feilke-2017-schreibundtextprozeduren}. Learning to write argumentative text is complex and requires continuous and detailed feedback \cite{kellogg-etal-2010-does,wambsganss-etal-2022-enhancing}.

Analyzing the argumentative structure of texts computationally, also known as argument(ation) mining, is a crucial and widely-studied step in providing automatic support for argumentative writing \cite{stede-schneider-2019-argumentation}. Student essays are a prominent domain for argument mining. A respective annotated corpus of 402 English student essays is available \cite{stab-gurevych-2017-parsing}, for which also quality issues such as insufficient claim support have been modeled \cite{stab-gurevych-2017-recognizing,gurcke:2021}. 
Additionally, student corpora are available for more specific domains, such as argumentative legal texts \cite{weber-etal-2023-structured}, persuasive peer reviews on business models \cite{wambsganss-etal-2020-corpus}, and business model pitches \cite{wambsganss-niklaus-2022-modeling}.

Some research has used argumentative structure to assess essay quality. In consecutive works, \citet{persing-etal-2010-organization} and \citet{persing-ng-2013-clarity, persing-ng-2014-prompt, persing-ng-2015-strength} graded different argumentation-related quality aspects for the well-known essay corpus ICLE \cite{granger-etal-2009-international}, namely organization, thesis clarity, prompt adherence, and argument strength. In contrast, \citet{horbach-etal-2017-fine} targeted different quality aspects of argumentative writing at once. Some works further investigated the interaction between argumentative structure and essay quality. \citet{wachsmuth-etal-2016-using} found that multiple argumentation-related essay scoring tasks benefit from argument mining, underlining the impact of argumentative structure on essay quality. The analyses by \citet{beigman-klebanov-etal-2016-argumentation} and \citet{nguyen-litman-2018-argument} suggest that this finding also holds for predicting holistic essay scores, while \citet{persing-etal-2010-organization} observed similar for the quality aspect organization. However, these studies relied on automatically assigned quality scores only, due to the lack of ground-truth annotations.

In a related line of research, approaches have been proposed to suggest revisions for argumentative essays \cite{afrin-litman-2018-annotation}, to assess the need for and the quality of revisions \cite{skitalinskaya:2023a,liu-etal-2023-predicting}, as well as to perform argument revisions computationally \cite{skitalinskaya-etal-2023-claim}. Other works towards writing support for argumentative essays presented a prototypical system that gives simple feedback in terms of missed criteria \cite{stab-2017-argumentative}, design principles for an adaptive learning tool \cite{wambsganss-rietsche-2019-towards}, visual feedback to the learner to prompt them to repair broken argument structures \cite{wambsganss-etal-2022-alen}, or point to enthymematic gaps in arguments and make suggestions on how to fill these gaps \cite{stahl-etal-2023-mind}. Most recently, \citet{britner-etal-2023-aquaplane} proposed a tool that not only analyzes issues with argument quality but also generates an explanation for its prediction. Our corpus supports these steps towards support systems for argumentative writing by providing detailed ground-truth annotations for both the argumentative structure and the quality of essays.

However, all the works mentioned deal with texts written by university students, while our work targets argumentative essays written by school students, fifth-graders and ninth-graders specifically. To the best of our knowledge, the only other published corpora with school student essays is not openly available \cite{correnti-etal-2013-assessing} and has been analyzed for essay-level quality aspects only, such as the integration of evidence \cite{rahimi-etal-2014-automatic} and the essay's organization \cite{rahimi-etal-2015-incorporating}. We recently came across another school student essay corpus in English with annotations for argumentative structure and quality, which has yet to be published.\footnote{The unpublished pre-print is available at \url{https://zenodo.org/records/8221504}.}
We go beyond in this work by assessing the quality of school student essays in terms of five aspects derived from language education literature while incorporating their interaction with annotated argumentative structures. This may foster the development of effective methods for helping school students improve their argumentative writing skills.

\section{School Student Essay Corpus}
\label{sec:scheme}

This section presents the source data and annotation of our corpus for analyzing the argumentative structure and quality of school student essays.

\subsection{Source Data}

As the basis, we systematically selected 1,320 German school student essays from the FD-LEX corpus \cite{becker-mrotzek-grabrowski-2018-textkorpus}.%
\footnote{Essays and metadata (grade, school form, age, gender, language background, and age group) are available here: \url{https://fd-lex.uni-koeln.de}}
The authors instructed students to each write three argumentative essays on topics pertinent to school students: 
(a) a letter to school funding organization on possible use of funding, 
(b) a statement on how to deal with the misbehavior of a fellow student, 
(c) a statement on who is guilty in a bike accident.

We seek to enable analyses of differences across different groups of school student essays on the corpus. Therefore, we pseudo-randomly chose 440 school students equally distributed across genders (only \emph{male} and \emph{female} exist in the corpus) and age groups (\emph{fifth-graders} and \emph{ninth-graders}). Subsequently, we included all three essays written by each selected school student from the source data.

\subsection{Annotation Scheme}

Our annotation scheme goes beyond existing corpora for argument mining, covering the macro and micro structure of argumentative essays on four levels in total. In addition, we evaluate the quality of the essays overall and in terms of four quality aspects. Figure~\ref{scheme} overviews our annotation scheme.

\hwfigure{scheme}{Proposed annotation scheme for argumentative school student essays: Four levels of argumentative macro and micro structure (discourse functions, arguments, components, discourse modes) and five essay quality aspects.}

\paragraph{Argumentative Structure}
On the broadest level of granularity for argumentative structure, we annotate {\it discourse functions} \cite{persing-etal-2010-organization}:
\begin{itemize}[itemsep=0pt, topsep=4pt]
	\item {\it Introduction.} Initiates an essay by presenting the topic and possibly the context of an essay. This section is usually non-argumentative and placed at the beginning of an essay.
	\item {\it Body.} Core of the essay, containing the majority of argumentative components.
	\item {\it Conclusion.} Summary of main points, often with a final evaluation of the topic. This section is typically found at the end of the essay.
\end{itemize}

\noindent
Next, we annotate {\it arguments} that comprise one point in an argumentative text, following \citet{walton-etal-2008-argumentation}. We differentiate them by stance towards the main standpoint (thesis) of an essay:
\begin{itemize}[itemsep=0pt, topsep=4pt]
	\item {\it Argument.} Ideally a claim (conclusion) and premises (reasons) supporting the claim.
	\item {\it Counter-argument.} An argument that attacks the thesis of an essay.
\end{itemize}

\noindent
For analyzing the micro structure, we annotate argumentative and non-argumentative {\it components}. As \citet{stab-gurevych-2017-parsing}, we also mark support and attack relations between them (see Figure~\ref{scheme}):
\begin{itemize}[itemsep=0pt, topsep=4pt]
    \item {\it Topic.} Non-argumentative component that describes the subject or purpose of the essay.
    \item {\it Thesis.} Main standpoint of the whole argumentative text towards the topic. Repetitions of the thesis are also annotated as such.
    \item {\it Antithesis.} Thesis contrary to the actual thesis.
   	\item {\it Modified Thesis.} Modified version of the actual thesis (e.g., more detailed or resticted).
    \item {\it Claim.} Statement that conveys a stance towards the topic.
    \item {\it Premise.} Reason that is given to support or attack a claim or another premise.%
    \footnote{The components in our annotation scheme are similar to those of \citet{stab-gurevych-2017-parsing}. We added {\it antithesis} and {\it modified thesis} to reflect the changes in position in the essays.}
\end{itemize}

\noindent
On the finest level of granularity, we annotate {\it discourse modes} \cite{smith-2003-modes} specific to school student essays. They are derived from language education literature, where they are used for developing and analyzing argumentative writing skills \cite{gaetje-etal-2012-positionierung, rezat-2018-argumentative, feilke-rezat-2021-textprozeduren}:
\begin{itemize}[itemsep=0pt, topsep=4pt]
	\item {\it Comparing.} Contrasting supporting and attacking points to a statement.
	\item {\it Conceding.} Addressing a counter-considera-tion and refuting it to support the own stance.
	\item {\it Concluding.} Drawing logical inferences using consecutive or final clauses (\emph{so that}, \emph{if... then}).
	\item {\it Describing.} Providing additional information, such as facts, statistics, and background data.
	\item {\it Exemplifying.} Providing examples or reporting on experiences.
	\item {\it Instructing.} Providing explicit instructions that recommend a specific course of action. 
	\item {\it Positioning.} Expressing the own standpoint.
	\item {\it Reasoning.} Providing causal links to support a claim/thesis using markers (\emph{because}, \emph{then}).
	\item {\it Referencing.} Mentioning statements made by others, for example, by authorities.
	\item {\it Qualifying.} Presenting a variation of the all-or-nothing standpoints.
\end{itemize}

\paragraph{Essay Quality}

As \citet{persing-etal-2010-organization}, we score essay quality on a 7-point scale from 1 (unsuccessful), 2 (rather unsuccessful), 3 (rather successful) to 4 (completely successful), with half points in between. We adapted the quality aspects of \citet{kruse-etal-2012-qualitat} for assessing school student essays in general to argumentative essays as follows:
\begin{itemize}[itemsep=0pt, topsep=4pt]
    \item {\it Relevance.} The essay fits the prompt.
    \item {\it Content.} The selection of content helps to reach the essay's goal.
    \item {\it Structure.} The selected points are coherent and well-connected.
    \item {\it Style.} The use of language is adequate.
    \item {\it Overall.} The overall impression of the rater. 
\end{itemize}

\subsection{Annotation Process}

We underwent the following process for both argumentative structure and essay quality:

To test and refine our annotation guidelines, we conducted pilot studies in which all annotators worked on the same 30 texts. We then discussed their understanding of the guidelines and annotation differences. We integrated their feedback into the guidelines to then test the reliability of the annotations in an inter-annotator agreement (IAA) study, where all annotators independently worked on the same 120 texts. Finally, each annotator annotated a set of 1,200 essays in the main annotation study.

As annotators, we employed experts in German language education from our lab and started the pilot and IAA studies on argumentative structure with three annotators. For the main part, only the two annotators with the most reliable annotations proceeded. The same annotators then annotated the essay quality, since they had already been trained in argumentative texts and our general procedure. However, we acknowledge that the annotators may have been predisposed to view essays in a certain way after the first annotation.

To assemble the final corpus, we combined the 1200 essays from the main study with the 120 essays from the IAA study after solving annotation conflicts. For conflicts between the three structure annotations per IAA essay, we kept the annotations that had the highest agreement across all levels with the other two. For conflicts between essay quality scores, we used their mean as the final score.\footnote{We rounded down to the next valid quality score for low values ($<2.5$) and rounded up for high values ($>2.5$) to prevent an upward distortion of the distribution.}

\subsection{Inter-Annotator Agreement}

\begin{table}[t]
    \centering
    \small
    \renewcommand{\arraystretch}{1.05}
    \setlength{\tabcolsep}{2.5pt}
    \begin{tabular}{lr}
        \toprule
        \bf Argumentative Structure &  \bf $\alpha$ \\
        \midrule
        Discourse Functions & 0.89 \\
        Arguments & 0.86 \\
        Components & 0.81 \\
        Discourse Modes & 0.74 \\
        Relations & 0.58 \\
        \bottomrule
    \end{tabular}
    \hspace{8pt}
    \begin{tabular}{lr}
        \toprule
        \bf Essay Quality &  \bf $\alpha$ \\
        \midrule
        Relevance & 0.77 \\
        Content & 0.95 \\
        Structure & 0.84 \\
        Style & 0.92 \\
        Overall & 0.95 \\
        \bottomrule
    \end{tabular}
    \caption{Krippendorff's $\alpha$ agreement in the IAA study between three annotators for argumentative structure and two annotators for essay quality. The high values stress the high reliability of our annotations.}
    \label{tab:iaa}
\end{table}

For the components, we follow \citet{stab-gurevych-2017-parsing} in that we evaluate the agreement per essay at the token level, so the token labels are the unit of analysis. Thereby, overlaps of annotations are taken into account. For relations, we determined the component-level spans that at least two annotators agreed on with a relative overlap $\geq$ 75\%. For all pairs of these, we then compared the relation labels (no relation, support, or attack) between the annotators. The mean Krippendorff's~$\alpha$ scores over the 120 IAA essays are reported in Table~\ref{tab:iaa}. For essay quality, we computed the $\alpha$-value per quality aspect with essays as the unit of analysis.

The agreement is high for argumentative structure spans with values between 0.74 and 0.89. The agreement for relations is lower but reasonable, given that disagreement from the components annotations is propagated to the relations. The agreement for essay quality is also high, too, ranging from 0.77 to 0.95. Overall, we conclude that the annotations can mostly be seen as very reliable. Content and style quality annotations are very consistent between annotators, while assessing the relevance and structure seems slightly more subjective.

\subsection{Corpus Statistics}

\begin{table}[t]
    \centering
    \small
    \renewcommand{\arraystretch}{1.05}
    \setlength{\tabcolsep}{1.75pt}
    \begin{tabular}{l@{\hspace*{-0.75em}}rrrr}
        \toprule
        \bf Label & \bf \# Spans & \bf \# Tokens & \bf Tokens/Span & \bf Spans/Essay  \\
        \midrule
		Introduction & 114 & 2\,329 & 20.43 & 0.09 \\
		Body & \bf 1\,335 & \bf 75\,766 & \bf 56.75 & \bf 1.01 \\
		Conclusion & 191 & 2\,938 & 15.38 & 0.14 \\
		\midrule
		Argument & \bf 2\,692 & \bf 51\,560 & \bf 19.15 & \bf 2.04 \\
		Counter-arg. & 34 & 514 & 15.12 & 0.03 \\
		\midrule
		Topic & 101 & 1\,656 & 16.40 & 0.08 \\
		Thesis & 1\,687 & 19\,581 & 11.61 & 1.28 \\
		Modified T. & 267 & 4\,490 & \bf 16.82 & 0.20 \\
		Antithesis & 14 & 174 & 12.43 & 0.01 \\
		Claim & \bf 3\,137 & \bf 39\,096 & 12.46 & \bf 2.38 \\
		Premise & 1\,020 & 12\,533 & 12.29 & 0.77 \\
		\midrule
		Comparing & 20 & 431 & \bf 21.55 & 0.02 \\
		Conceding  & 142 & 2\,874 & 20.24 & 0.11 \\
		Concluding  & 868 & 11\,654 & 13.43 & 0.66 \\
		Describing  & 1\,692 & \bf 22\,258 & 13.15 & 1.28\\
		Exemplifying & 63 & 926 & 14.70 & 0.05 \\
		Instructing  & 176 & 2\,174 & 12.35 & 0.13\\
		Positioning & \bf 1\,758 & 19\,178 & 10.91 & \bf 1.33 \\
		Reasoning  & 1\,553 & 17\,204 & 11.08 & 1.18 \\
		Referencing  & 16 & 197 & 12.31 & 0.01 \\
		Qualifying & 147 & 2\,344 & 15.95 & 0.11 \\
        \bottomrule
    \end{tabular}
    \caption{Argumentative structure annotations in the corpus: Total number of spans and tokens per label, average span length in number of tokens ({\it Tokens/Span}) and average number of spans per essay ({\it Spans/Essay}). The highest value per column and level is marked bold.}
    \label{tab:sequence-stats}
\end{table}

Table~\ref{tab:sequence-stats} gives insights into the label distribution for argumentation structure. \emph{Body} occurs most frequently among the discourse functions (1,335 times). With 56.75 tokens on average, bodies are also notably longer than \emph{introductions} and \emph{conclusions}. On the argument level, we note that \emph{counter-arguments} are rather sparse in student essays. Among the components, \emph{claims} are most frequent in total and per essay, followed by \emph{theses}. Furthermore, we notice that \emph{modified theses} are usually longer than theses, which matches our expectation that students add more details or restrictions to the thesis there. The most used discourse modes are \emph{positioning}, \emph{describing}, and \emph{reasoning}, while \emph{referencing}, \emph{comparing}, and \emph{exemplifying} occur rarely. Notable are also the differences in span length, e.g. \emph{positioning} spans have on average only about half as many tokens as \emph{comparing} spans.

\begin{table}[t]
    \centering
    \small
    \renewcommand{\arraystretch}{1.05}
    \setlength{\tabcolsep}{2pt}
    \begin{tabular}{lrr}
        \toprule
        \bf From \textbf{\textit{Claim}} to & \bf \# & \bf \% \\ 
        \midrule
		Thesis & \bf 2844 & \bf 92.6 \\
		Modified Thesis & 218 & 7.1 \\
		Antithesis & 9 & 0.3 \\	 
        \bottomrule
    \end{tabular}
	\hspace{5pt}
	\begin{tabular}{lrr}
		\toprule
		\bf From \textbf{\textit{Premise}} to & \bf \# & \bf \% \\ 
		\midrule
		Thesis & 6 & 0.6 \\
		Claim & \bf 872 & \bf 85.8 \\
		Premise & 138 & 13.6 \\ 		 
		\bottomrule
	\end{tabular}
    \caption{Absolute and relative frequency of annotated relations outgoing from {\it claim} (left) or {\it premise} (right).}
    \label{tab:relation-stats}
\end{table}

Table~\ref{tab:relation-stats} gives the frequency of annotated relations in our corpus. Most relations outgoing from {\it claims} are directed towards {\it theses} (92.6\%), while most relations outgoing from {\it premises} are directed towards {\it claims} (85.8\%). Overall, 96.2\% of the relations were labeled as {\it support} and 3.8\% as {\it attack}.

\begin{table}[t]
    \centering
    \small
    \renewcommand{\arraystretch}{1.05}
    \setlength{\tabcolsep}{2.6pt}
    \begin{tabular}{lrrrrrrrr}
        \toprule
        \bf Quality Aspect & \bf 1.0 & \bf 1.5 & \bf 2.0 & \bf 2.5 & \bf 3.0 & \bf 3.5 & \bf 4.0 & \bf Mean \\
        \midrule
		Relevance & 64 & 131 &  \bf 548 & 358 & 190 & 23 & 6 & 2.22 \\
		Content & 62 & 33 & 123 & 312 &  \bf 510 &  \bf 173 & \bf  107 &  \bf 2.80 \\
		Structure & 59 & 91 & 423 & 414 & 292 & 24 & 17 & 2.35 \\
		Style & 75 &  \bf 159 & 396 &  \bf 436 & 233 & 12 & 9 & 2.25 \\
		Overall &  \bf 90 & 142 &478 & 390 & 195 & 23 & 2 & 2.20 \\
        \bottomrule
    \end{tabular}
    \caption{Distribution and mean of scores per quality aspect. The highest value per column is marked bold.}
    \label{tab:rating-stats}
\end{table}

The distribution of quality scores is shown in Table~\ref{tab:rating-stats}. We can see that \emph{relevance}, \emph{structure}, and \emph{style} have a similar score distribution, while the distribution of \emph{content} scores is shifted towards the higher scores, with the highest mean (2.80). Overall quality has the lowest mean score (2.20) and was most often scored with the lowest score of 1.0. These results suggest that overall quality is not just the average of the other annotated quality aspects but that it emerges from the annotators' perception and possibly other aspects.

\section{Analysis}
\label{sec:analysis}

\bsfigure{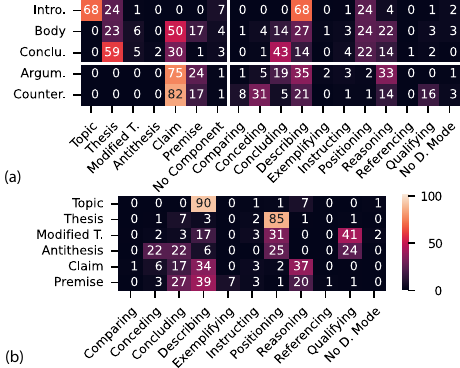}{Cooccurrence matrices: Relative token-level overlap of (a)~macro and micro structure and (b)~component and discourse mode labels in percent. For example, 68\% of all tokens labeled as {\it Introduction} on the macro level are also labeled as {\it Topic} on the micro level.}

This section reports on our corpus analysis of the interaction between the argumentative structure on micro vs.\ macro level and component vs.\ discourse mode level, and the different essay quality aspects.

\subsection{Macro vs.\ Micro structure}

Figure~\ref{heatmaps}(a) shows the overlap between macro structure (discourse functions and arguments) and micro structure (components and discourse mode) labels. The \emph{introduction} mainly includes the \emph{topic} (68\%), while every second token in the \emph{body} is a \emph{claim} on average. The \emph{thesis} cooccurs with all three discourse functions. We see that the proportion of \emph{claims} and \emph{premises} differs for \emph{arguments} and \emph{counter-arguments}. Counter-arguments contain more claim tokens than arguments, while fewer counter-argument tokens are part of a premise.

The usage of discourse modes differs between the macro-structure levels, too. In the introduction, students mostly \emph{describe} (68\%) and \emph{position} (24\%). The discourse modes are more diverse for the body, also including a notable portion of \emph{reasoning}. As expected, in the \emph{conclusion}, students focus more on \emph{concluding}. While describing and reasoning are prevalent in arguments and counter-arguments, a notable portion of argument tokens is used for concluding. At the same time, more \emph{conceding} and \emph{qualifying} tokens occur in counter-arguments. This is expected, since especially in counter-arguments other points of view should be varied or refuted.

\subsection{Components vs.\ Discourse modes}

The cooccurrences between components and actions can be seen in Figure~\ref{heatmaps}(b). While the \emph{topic} is mostly \emph{described} (90\%) and the \emph{thesis} consists primarily of \emph{positioning} (85\%), the remaining components include more diverse discourse modes. In contrast to \emph{theses}, \emph{modified theses} also feature \emph{describing} and \emph{qualifying} tokens, while \emph{antitheses} additionally cover \emph{conceding} and \emph{concluding}. \emph{Claims} and \emph{premises} mainly cooccur with \emph{describing}, \emph{reasoning}, and \emph{concluding}. However, the proportions differ slightly. The cooccurrence matrix between all structure labels can be found in Appendix~\ref{app:cooccurrence}.

\subsection{Essay Quality}

\begin{table}[t]
    \centering
    \small
    \renewcommand{\arraystretch}{1.05}
    \setlength{\tabcolsep}{2.2pt}
    \begin{tabular}{lccccc}
        \toprule
         & \bf Relevance & \bf Content & \bf Structure & \bf Style & \bf Overall \\
        \midrule
		Relevance & & .53 & .61 & .47 & \bf .75 \\
		Content & .53 & & .48 & .41 & .60 \\
		Structure & .61 & .48 & & .51 & .71 \\
		Style & .47 & .41 & .51 & & .61 \\
		Overall & \bf .75 & \bf .60 & \bf .71 & \bf .61 \\
        \bottomrule
    \end{tabular}
    \caption{Kendall’s $\tau$ correlation between the quality aspects. The highest value per column is marked bold.}
    \label{tab:rating-correlation}
\end{table}

To further assess the interaction between the quality aspects, Table~\ref{tab:rating-correlation} shows all pairwise Kendall's~$\tau$ correlations. All aspects correlate most with \emph{overall quality}, most strongly \emph{relevance} (.75). The correlation between \emph{content} and \emph{style} is lowest (.41), which underlines their distinctive nature.

\section{Experiments}
\label{sec:experiments}

This section presents baseline approaches to the two main tasks our corpus enables: Predicting the argumentative structure (argument mining) and the essay quality (essay scoring). Additionally, we investigate whether information about the argumentative structure helps to predict the essay quality.

\subsection{Argument Mining}

We treat argument mining as a token classification task: Given a school student essay and a structure level, predict the label of each token on that structure level. The IOB2 format is used for the labels to separate adjacent spans of the same type. We performed 5-fold cross-validation for each structure level. For each folding, we used four folds (80\%) for training and divided the fifth fold in half: one half (10\%) for selecting the best-performing checkpoint in terms of macro-averaged F$_1$-score, and the remaining half (10\%) for testing.

\paragraph{Models} 

We used the multilingual model \emph{mDeBERTaV3} \cite{he-etal-2023-debertav3} ({microsoft/mdeberta-v3-base}) from Huggingface \cite{wolf-etal-2020-transformers}.%
\footnote{Explorative experiments using instruction fine-tuned models such as Alpaca \cite{taori-etal-2023-standford} did not lead to promising results for our token classification task.}
Besides, we tested the effect of training \emph{adapters} \cite{houlsby-etal-2019-parameter}, a set of task-specific parameters that are added to every transformer layer of mDeBERTaV3 and fine-tuned on the task while the model weights are fixed. To quantify the impact of learning, we compare against a random baseline that chooses a token label pseudo-randomly and a majority baseline that always predicts the majority token label from the training set. As upper bound, we report the human performance in terms of the average of each annotator in isolation on the 120 IAA texts annotated by all annotators.\footnote{Note that the test set used for the model performance is a different subset of the dataset.} 

\begin{table}[t]
	\centering
	\small
	\renewcommand{\arraystretch}{1.05}
	\setlength{\tabcolsep}{2pt}
	\begin{tabular}{lrrrrrrrrrrr}
		\toprule
		& \multicolumn{2}{c}{\bf D. Func.} && \multicolumn{2}{c}{\bf Argum.} && \multicolumn{2}{c}{\bf Compon.} && \multicolumn{2}{c}{\bf D. Mode}  \\
		\cmidrule{2-3} \cmidrule{5-6} \cmidrule{8-9} \cmidrule{11-12}
		\bf Approach & \bf Acc. & \bf F$_1$ && \bf Acc. & \bf F$_1$ && \bf Acc. & \bf F$_1$ && \bf Acc. & \bf F$_1$  \\
		\midrule
		Random 		& .14 & .00 && .20 & .00 && .08 & .00 && .05 & .00 \\ 
		Majority 	& .86 & .52 && .56 & .00 && .41 & .00 && .24 & .00 \\
		\addlinespace
		mDeBERTaV3 	& .92 & .46 && .86 & .29 && .66 & .21 && .63 & .21 \\
		-adapter  & \bf .95 & \bf .68 && \bf .92 & \bf .52 && \bf .76 & \bf .49 && \bf .73 & \bf .46 \\
		\midrule
		Human & .98 & .94 && .96 & .85 && .93 & .89 && .89 & .84 \\
		\bottomrule
	\end{tabular}
	\caption{Argument mining results: Macro F$_1$-score and accuracy of each approach in 5-fold cross-validation on all four argumentative structure dimensions. The best value per column is marked bold.}
	\label{tab:argmining-results}
\end{table}

\begin{table*}
	\centering
	\small
	\renewcommand{\arraystretch}{1.1}
	\newcolumntype{R}[1]{>{\raggedleft\let\newline\\\arraybackslash\hspace{0pt}}m{#1}}
	\setlength{\tabcolsep}{3.5pt}
	\begin{tabular}{lR{2.1cm}R{2.1cm}R{2.1cm}R{2.1cm}R{2.1cm}}
		\toprule
 		\bf Approach & \multicolumn{1}{c}{\bf Relevance} & \multicolumn{1}{c}{\bf Content} & \multicolumn{1}{c}{\bf Structure} & \multicolumn{1}{c}{\bf Style} & \multicolumn{1}{c}{\bf Overall} \\
		\midrule
		Random 		& $-$0.013 \scriptsize $\pm 0.084$ & $-$0.011 \scriptsize $\pm 0.071$ & $-$0.014 \scriptsize $\pm 0.073$ & 0.017 \scriptsize $\pm 0.084$ & $-$0.004 \scriptsize $\pm 0.083$ \\ 
		Majority 	& 0.000 \scriptsize $\pm 0.000$ & 0.000 \scriptsize $\pm 0.000$ & 0.000 \scriptsize $\pm 0.000$ & 0.000 \scriptsize $\pm 0.000$ & 0.000 \scriptsize $\pm 0.000$ \\
		mDeBERTaV3 	& 0.530 \scriptsize $\pm 0.069$ & 0.295 \scriptsize $\pm 0.109$ & 0.513 \scriptsize $\pm 0.044$ & 0.492 \scriptsize $\pm 0.059$ & 0.616 \scriptsize $\pm 0.040$ \\
		-adapter & 0.564 \scriptsize $\pm 0.018$ &  0.431 \scriptsize $\pm 0.098$ &  \bf 0.575 \scriptsize $\pm 0.038$ & 0.579 \scriptsize $\pm 0.077$ & 0.648 \scriptsize $\pm 0.054$ \\
		
		\addlinespace
		
		-fusion-w/-discourse-functions & 0.599 \scriptsize $\pm 0.043$ & 0.381 \scriptsize $\pm 0.134$ & 0.559 \scriptsize $\pm 0.036$ & 0.569 \scriptsize $\pm 0.069$ & 0.668 \scriptsize $\pm 0.049$ \\
		-fusion-w/-arguments & 0.593 \scriptsize $\pm 0.030$ & 0.448 \scriptsize $\pm 0.105$ & \bf 0.575 \scriptsize $\pm 0.019$ & 0.581 \scriptsize $\pm 0.054$ & 0.668 \scriptsize $\pm 0.036$ \\			
		-fusion-w/-components & \bf \textsuperscript{\textdagger}0.600 \scriptsize $\pm 0.025$ & 0.437 \scriptsize $\pm 0.137$ & 0.543 \scriptsize $\pm 0.044$ & 0.585 \scriptsize $\pm 0.053$ & 0.663 \scriptsize $\pm 0.046$ \\
		-fusion-w/-discourse-modes & 0.544 \scriptsize $\pm 0.028$ & 0.420 \scriptsize $\pm 0.118$ & 0.535 \scriptsize $\pm 0.041$ & 0.583 \scriptsize $\pm 0.064$ & 0.645 \scriptsize $\pm 0.023$ \\			
		-fusion-w/-all & 0.574 \scriptsize $\pm 0.039$ & \bf 0.454 \scriptsize $\pm 0.142$ & 0.546 \scriptsize $\pm 0.013$ & \bf 0.617 \scriptsize $\pm 0.057$ & \bf \textsuperscript{\textdagger}0.686 \scriptsize $\pm 0.031$ \\
		\midrule
		Human & 0.636 \scriptsize $\pm 0.055$ & 0.632 \scriptsize $\pm 0.003$ & 0.734 \scriptsize $\pm 0.007$ & 0.766 \scriptsize $\pm 0.005$ & 0.746 \scriptsize $\pm 0.003$ \\
		\bottomrule
	\end{tabular}
	\caption{Essay scoring results: QWK of each approach in 5-fold cross-validation on all five quality dimensions. The best value per column is marked bold. We mark significant gains over {\it mDeBERTaV3-adapter} at $p < .05$ with $\dag$.}
	\label{tab:quality-results}
\end{table*}

\paragraph{Experimental Setup} 

We train mDeBERTaV3 for 30 epochs (1,980 steps) using the suggested hyperparameter values: a learning rate of $3e-5$, batch size 16, and 500 warmup steps. For mDeBERTaV3-adapter, we follow \citet{pfeiffer-etal-2020-adapterhub} who recommend to use a higher learning rate of $1e-4$ and train longer, here 50 epochs (3,300 steps).

\paragraph{Results} 

Table~\ref{tab:argmining-results} presents the token classification results for all levels of argumentative structure, averaged over all folds. Noteworthily, \emph{mDeBERTaV3-adapter} outperforms training the whole model (\emph{mDeBERTaV3}) in all cases. Given that the F$_1$-scores improve more than the accuracy, the adapters seem less prone to overfitting to the majority label. This learning success suggests the possibility of predicting all argumentative structure levels on our corpus. However, further improvements using more advanced approaches are expected.

\subsection{Essay Scoring}

We treat predicting the essay quality as a text classification task: Given a school student essay and a quality aspect, predict the corresponding quality score. As before, we performed 5-fold cross-validation for each quality aspect using the same folds. We selected the best-performing checkpoint on the validation set using quadratic weighted kappa (QWK), the most widely adopted metric for automatic essay scoring \cite{ke-ng-2019-automated}. 

\paragraph{Models} 

We adopted the previous approaches by changing the head to a text classification head. To analyze the interaction between argumentative structure and essay quality, we employed \emph{AdapterFusion} \cite{pfeiffer-etal-2021-adapterfusion}, a multi-task learning framework that can be used to investigate relations between different dimensions by learning how to combine model weights with one or more adapters. We used the mDeBERTaV3-adapters trained on argumentative structure from the previous experiment. As the final adapter, we chose the one trained on the folding that performed most representative for all folds (F$_1$-score closest to the reported averaged F$_1$-score across folds). To measure the impact of each level of argumentative structure on the scoring performance, we used each adapter individually and a combination of all of them.

\paragraph{Experimental Setup} 

The experimental setup for mDeBERTaV3 and mDeBERTaV3-adapter was adopted from before. For training the AdapterFusion, we followed \citet{pfeiffer-etal-2021-adapterfusion} to use a learning rate of $5e-5$ and trained shorter than the adapters, in our case for 20 epochs (1,320 steps).

\paragraph{Results} 

Table~\ref{tab:quality-results} shows the scoring results. All models outperform the lower-bound baselines (\emph{random} and \emph{majority}), suggesting that the quality scoring can be learned from our corpus. Furthermore, fusing all adapters trained on argumentative structure ({\it mDeBERTaV3-fusion-w/-all}) performs best for three out of five quality aspects, significantly beating mDeBERTaV3-adapter in predicting overall quality.\footnote{We use Wilcoxon signed-rank tests at $p < .05$ for testing the significance.} This underlines the need for all four levels of argumentative structure together in order to improve scoring overall quality (0.686 vs.\ 0.648). In addition, using only the adapter trained on the component level ({\it mDeBERTaV3-fusion-w/-components}) helps to significantly improve over mDeBERTaV3-adapter in predicting relevance (0.600 vs.\ 0.564), indicating an interaction between the structure on component level and this quality aspect. QWK scores greater or equal to 0.6 suggest substantial agreement between the predicted and ground-truth quality scoring of essays.

\hwfigure{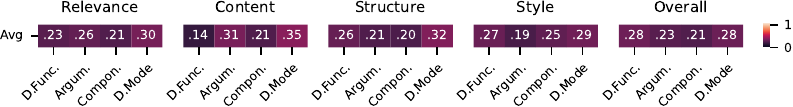}{AdapterFusion activation on average over the layers for each {\it mDeBERTaV3-fusion-w/-all} model per quality aspect. We average the activation for each fused adapter (discourse functions, arguments, components, discourse modes) over all instances in the most representative test set folding.}

\paragraph{AdapterFusion Activations} 

AdapterFusion extracts information from adapters only if they benefit the target task. 
%
Similar to \citet{falk-lapesa-2023-bridging}, we visualize the average activations of our model {\it mDeBERTaV3-fusion-w/-all} over the layers in Figure~\ref{activation-avg} to investigate the influence of each level of argumentative structure on the quality scoring. All adapters are activated fairly evenly for all quality aspects, with slight deviations. This aligns with our previous results and underlines that all annotated structure levels are helpful for quality scoring. The activations per layer can be found in Appendix \ref{app:fusion-activation}.

\section{Conclusion}
\label{sec:conclusion}

Argumentative writing support of school students presupposes that the quality of their arguments can be assessed. Until now, no argument mining corpus with school student essays has been published, let alone any essay corpus with both argument and quality annotations. With this work, we fill both research gaps with a new corpus of 1,320 German school student essays, annotated by experts for argumentative structure and essay quality.

Our corpus analysis has provided various insights into the correlation between the different levels of argumentative structure and essay quality. In our experiments with fine-tuned transformers and adapters for mining argumentative structure and scoring essay quality we have demonstrated that combining information on all four argumentative structure levels helps the prediction of essay quality. This shows the usefulness of our corpus for research on quality-oriented argumentative writing support, which we seek to enable with this paper.

We point out that our corpus contains various information yet to be explored, such as argumentative relations and school student metadata. It thus lays the ground for further analyses---like identifying unwarranted claims and studying differences across age groups and genders.

\section{Limitations}

Aside from the still-improvable performance of the presented baseline models for argument mining and essay scoring, we see two notable limitations of our work: the restriction to German texts, and the pending utilization of the corpus for quality-oriented argumentative writing support.

First, we point out the specific language background of our work. The essays were written by German school students, and the annotations were developed in close communication with German experts from the field of language education, while the discourse modes and essay quality aspects are, to a considerable extent, derived from work on German texts. This means that our findings may not perfectly align with argumentative writing in other countries or languages with different expectations for argumentative essays. 

Second, while our analyses suggest that our corpus helps to enable quality-oriented argumentative writing support, the perceived usefulness of such a tool is still to be evaluated. We expect and encourage future work to utilize our corpus for such writing support tools, for example, by further analyzing which exact argumentative structures influence the essay quality and to what extent. Interpretable essay quality scoring based on the structure might generate helpful insights that can be used as writing feedback by school students.

\section{Ethical Considerations}

We see no apparent risk of the corpus or the methods presented in this paper being misused for ethically doubtful purposes. The authors of the FD-LEX corpus \cite{becker-mrotzek-grabrowski-2018-textkorpus} have already pseudonymized the author of each essay. Therefore, it is not possible to identify the individual school student from the provided data. However, we want to point out that one might find differences in the essays across gender or age groups that do not reflect reality but are rather due to an unintentional bias in the data selection.

\section*{Acknowledgments}
We would like to thank the participants of our study and the anonymous reviewers for the valuable feedback and their time. This work was partially funded by the Deutsche Forschungsgemeinschaft (DFG, German Research Foundation) within the project ArgSchool, project number 453073654.

\bibliography{naacl24-learner-corpus-lit}
\bibliographystyle{acl_natbib}

\appendix
\section{Cooccurrence Matrix}
\label{app:cooccurrence}

\hwfigure{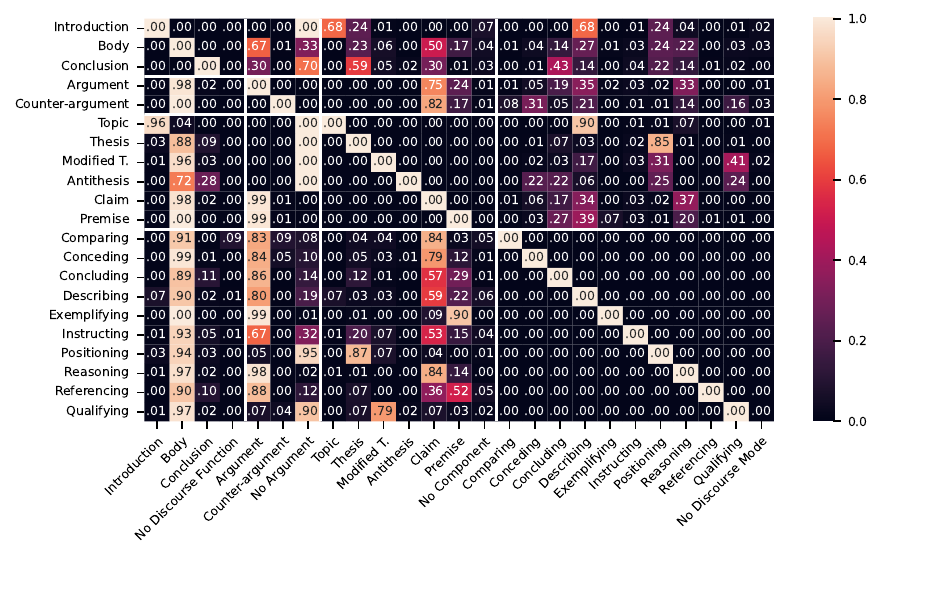}{Relative token-level overlap of all argumentative structure labels, seperated into the four levels of granularity. For example, 68\% of all tokens labeled as {\it Introduction} are also labeled {\it Topic}.}

The cooccurrence matrix between all argumentative structure levels (discourse functions, arguments, components, and discourse modes) is shown in Figure~\ref{heatmap-cooccurrences}.

\section{AdapterFusion Activations}
\label{app:fusion-activation}

\hwfigure{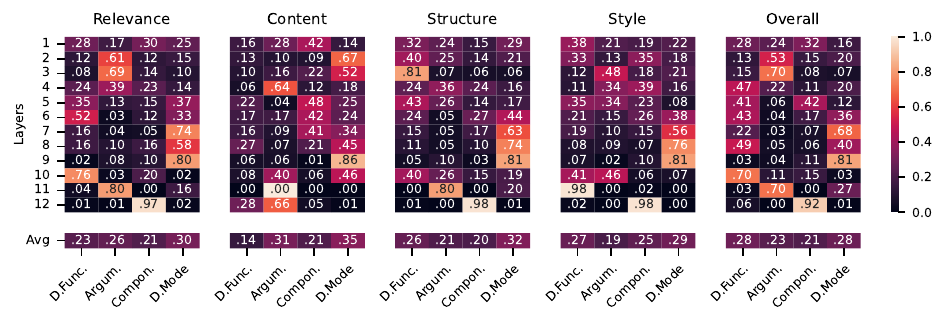}{AdapterFusion activation per layer ({\it 1--12}) and on average over the layers ({\it Avg}) for each {\it mDeBERTaV3-fusion-w/-all} model per quality aspect. We average the activation for each fused adapter (for discourse functions, arguments, components, or discourse modes) over all instances in the test set of the most representative folding.}

Similar to \citet{pfeiffer-etal-2021-adapterfusion}, we visualize the activations of our model {\it mDeBERTaV3-fusion-w/-all} per layer in Figure~\ref{activation-viz} to further investigate the influence of each level of argumentative structure on the quality scoring. The first activation layers show for all five quality aspects that all structure adapters are activated quite diversely. In contrast, the later layers have a clear tendency towards activating only one or two adapters. Notable is the similar activation pattern between relevance and overall quality, which could come from their value correlation.

\end{document}